\documentclass[12pt,onecolumn]{IEEEtran}

\usepackage{floatflt}
\usepackage{multirow}
\usepackage{hyperref}
\usepackage{graphicx}
\usepackage{url}
\usepackage{color}
\usepackage{subfigure}
\usepackage{amssymb}
\usepackage{amsmath}
\usepackage{times}
\usepackage[pagewise]{lineno}

\def\a{\mathbf a}
\def\b{\mathbf b}

\def\p{\mathbf p}

\def\r{\mathbf r}

\def\t{\mathbf t}

\def\v{\mathbf v}

\def\x{\mathbf x}
\def\y{\mathbf y}
\def\z{\mathbf z}

\newtheorem{algorithm}{Algorithm}

\newtheorem{theorem}{Theorem}

\title{A Model for \\
Auto-Programming for General Purposes}

\author{
\IEEEauthorblockN{Juyang Weng\IEEEauthorrefmark{1}\IEEEauthorrefmark{2}\IEEEauthorrefmark{3}\\
\IEEEauthorblockA{\IEEEauthorrefmark{1}Department of Computer Science and Engineering,\\ 
\IEEEauthorrefmark{2}Cognitive Science Program,\\
\IEEEauthorrefmark{3} Neuroscience Program, \\
Michigan State University, East Lansing, MI, 48824 USA}}
}

\begin{document}
\maketitle



{\bf Abstract:}  The Universal Turing Machine (TM) is a model for VonNeumann computers --- general-purpose computers.   A human brain can inside-skull-automatically learn a  universal TM so that he acts as a general-purpose computer and writes a computer program for any practical purposes.  It is unknown whether a machine can accomplish the same.  This theoretical work shows how the Developmental Network (DN) can accomplish this.   Unlike a traditional TM, the TM learned by DN is a super TM --- Grounded, Emergent, Natural, Incremental, Skulled, Attentive, Motivated, and Abstractive (GENISAMA).  A DN is free of any central controller (e.g., Master Map, convolution, or error back-propagation).  Its learning from a teacher TM is one transition observation at a time, immediate, and error-free until all its neurons have been initialized by early observed teacher transitions.  From that point on, the DN is no longer error-free but is always optimal at every time instance in the sense of maximal likelihood, conditioned on its limited computational resources and the learning experience.  This letter also extends the Church-Turing thesis to automatic programming for general purposes and sketchily proved it.

%
%



%
%
%
\newpage
\setcounter{page}{2}
It remains elusive how a biological brain represents, computes, learns, memorizes, updates, and abstracts through its life-long experience---from a zygote, to embryo, fetus, newborn, infancy, childhood, and adulthood.
Gradually, the brain produces behaviors that are increasingly rule-like \cite{Piaget54,Cole96,Sur05} and can perform auto-programming for general purposes.  By auto-programming, we mean that a brain automatically generates a sequence of procedures, from tying shoelaces, to making a business plan, to writing a computer program.    Such programs are not just random shufflers.  They must relate to meanings of the world --- namely physics gives rise to meanings \cite{Harnad90,Muller10}.

Here, we greatly simplify such rich processes of co-development of brain and body through activities, 
assisted by innate (i.e., prenatally developed) reflexes and  innate motivations \cite{Blakemore70,LiFitzpatrick06}, to realize auto-programming from facts, education, engineering, thinking, fiction, and discovery. We ask only:  What
is a minimal set of mechanisms that enables a biological or silicon machine to learn auto-programming for general purposes?  Some early examples from an answer below are in a companion paper. 

Three conceptual steps guide us toward this answer.  We first extend Finite Automata (FAs) \cite{Hopcroft06,Martin11} to agents in the sense that states are not hidden but are open as actions.  Then we extend such agent FAs to attentive agent FAs, so that the machines can automatically attend 
only a subset of current inputs (e.g., some words among all words on this page).  Finally we introduce the GENISAMA TM by replacing all symbols in such attentive agent FAs with patterns that naturally emerge from the real world.  

{\bf Agent FA:}  Two variants of FA, Moore machines and Mealy machines \cite{Hopcroft06,Martin11} output actions but not their states.  We extend an FA to agent \cite{Russell10}, called {\em Agent FA}, by simply requiring it to output its current state entirely, but its current actions are included in the current state.   This extension is conceptually important because the current state is now teachable as actions so that we are ready to address the issue of internal representations in neural networks below.   In psychology, 
all skills and knowledge fall into two categories \cite{Sun05}, declarative (e.g., verbal) and non-declarative (e.g., bike riding).   Therefore, all skills and knowledge can be expressed as actions.

{\bf Attentive Agent FA:}  Suppose that a symbolic street scene at time $t$ has multiple objects.  E.g.,  
\[ 
S(t)=\{\mbox{car1, car2, sign1, sign2, pedestrian1, ...} \}
\]
Instead of taking only one input symbol $\sigma$ at a time (e.g., $\sigma=\mbox{car1}$),  an attentive agent FA attends to a set of symbols at a time (e.g., $s(t)=\{\mbox{car1, pedestrian1\}}\subset S(t)$).    The control of any TM is an Attentive Agent FA as we will discuss below.  

\begin{figure}[tb]
     \centering
      \includegraphics[width=4.5in]{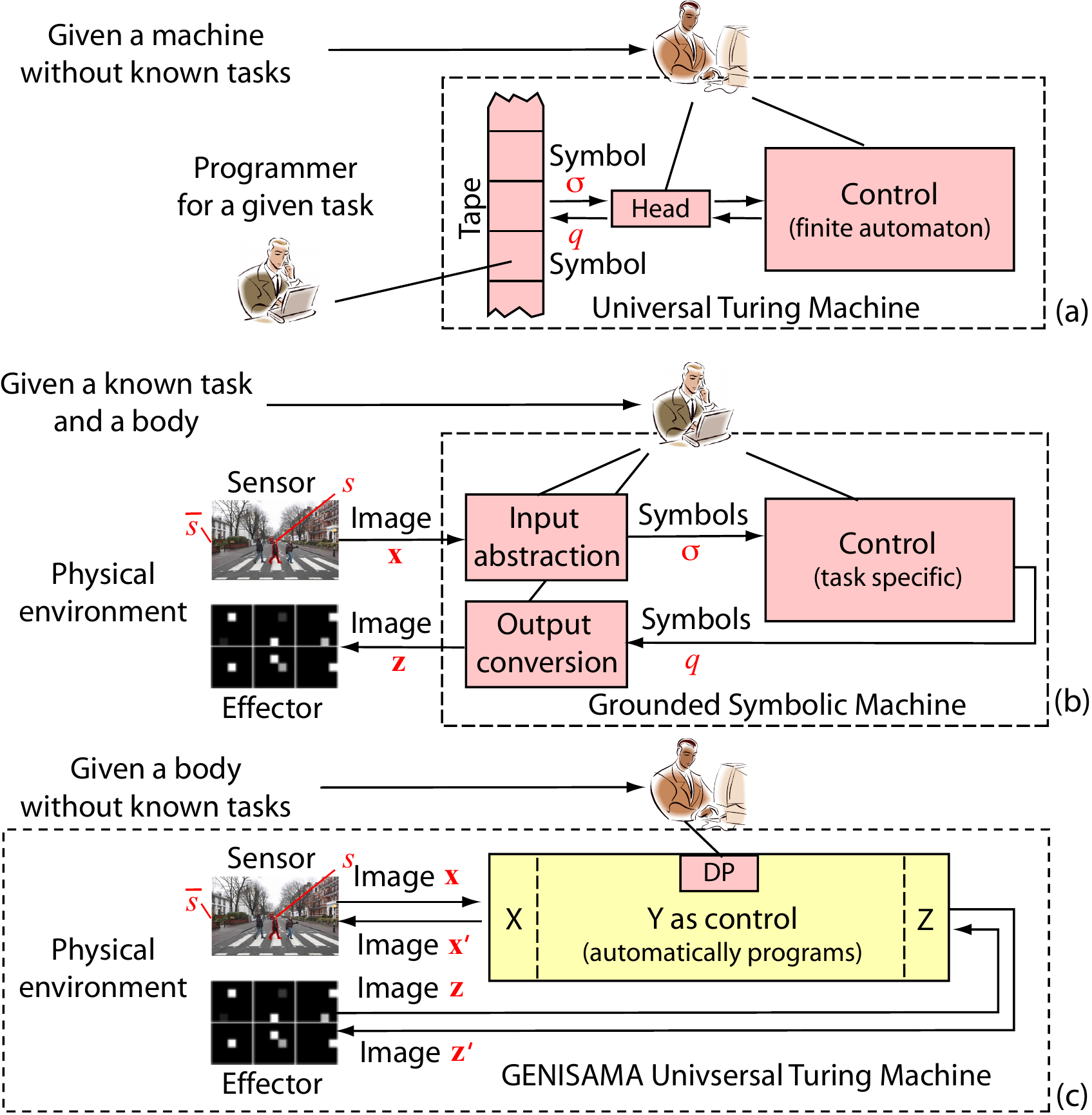}
\caption{\protect\small Three categories of agents, (a) Universal Turing Machines that is symbolic and cannot auto-program, (b) Grounded Symbolic Machines that are task-specific and cannot auto-program, and (c) GENISAMA Universal Turing Machines whose Developmental Program are task-nonspecific. A GENISAMA Universal Turing Machines can auto-program for general purposes.  The tape in (a) becomes the real world and all the symbols in (a) and (b) become natural patterns.  DP: Developmental Program.  $X$: the sensory port.  $Z$: the effector port.  $Y$:  the hidden
``bridge for ``banks'' $X$ and $Z$.   Pink block: human handcrafted. Yellow blocks: emerge automatically.}
\label{FG:TM-GSM-GENISAMA}
\end{figure}

In order to understand auto-programming for general purposes, we need to first discuss the Universal TM \cite{Turing,Hopcroft06,Martin11}.     

Recently, it has been proved \cite{WengIJIS15} that the control of any TM is an FA as illustrated in Fig.~\ref{FG:TM-GSM-GENISAMA}(a).  Using this new result, our examples in Methods are much simpler. 

\begin{theorem}
\label{TM:AttentiveFA}
The control of a TM is not only an Agent FA, but also an Attentive Agent FA. 
\end{theorem}

The proof is in Methods.

A Universal TM is for general purposes \cite{Hopcroft06,Martin11}.  
The input tape of a Universal TM has two parts, the program as instructions and the data for the program to use, not just data like a regular TM. 
Theorem~\ref{TM:AttentiveFA} is also true for any Universal TM because it is a special kind of TM. 
 
Because the input is a set of symbols instead of a symbol, the transition table of an Attentive 
Agent FA, especially as the control of a Universal TM, is typically extremely large --- impractical to handcraft.      

\begin{figure*}[p]
     \centering
      \includegraphics[width=5.5in]{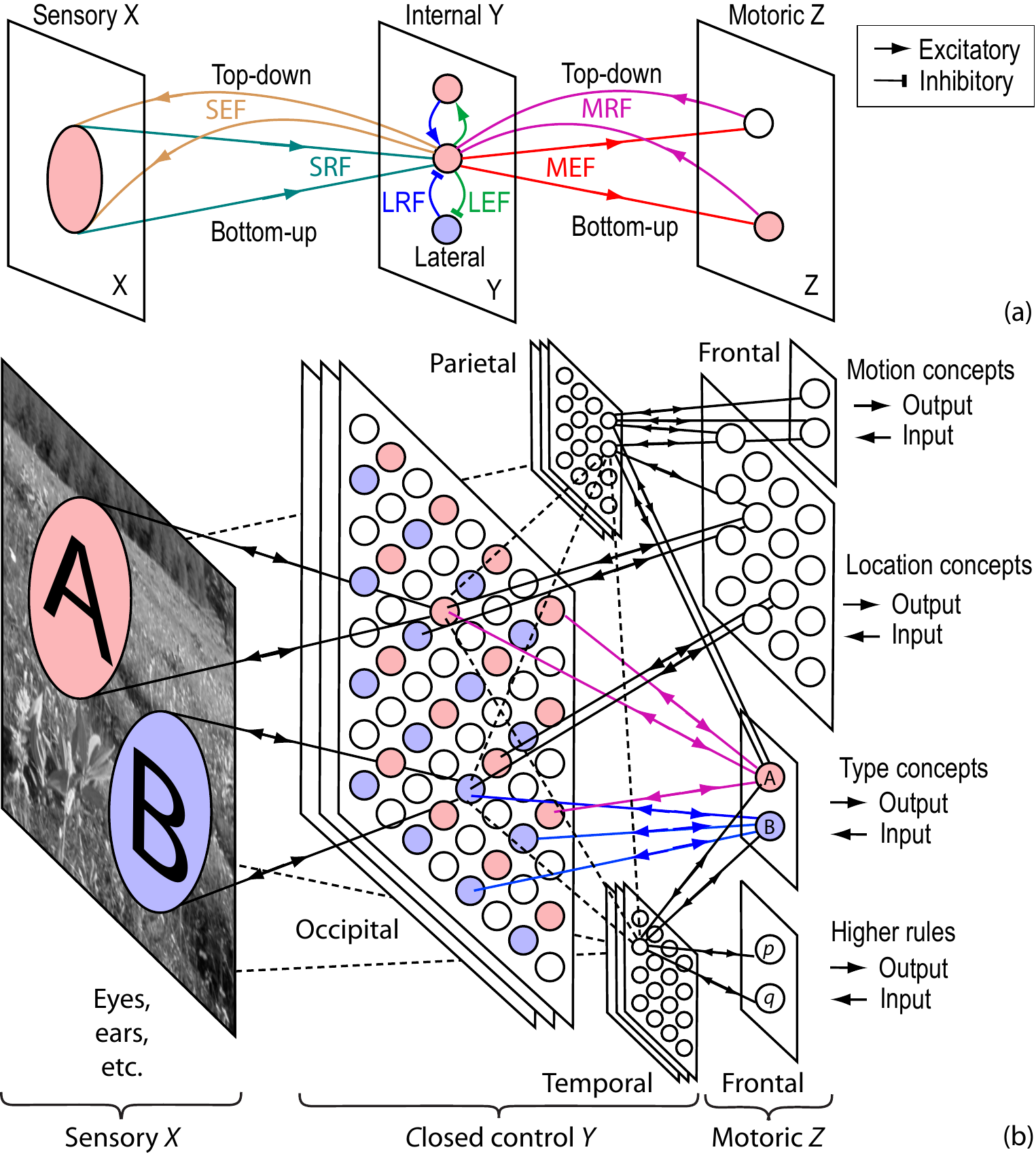}
\caption{\protect\small Brain $Y$ is theoretically modeled as the two-way bridge of the sensory bank $X$ 
and the motor bank $Z$. The bridge, mathematically denoted by Eq.~\eqref{EQ:DNtransition}, is
extremely rich: Self-wiring within a Developmental Network (DN) as the control of GENISAMA TM, based on statistics of activities through ``lifetime'', without any central controller,  Master Map, handcrafted features, and convolution.  (a) Each feature neuron has 
six fields in general. S: Sensory; M: motoric; L: lateral; R: receptive; E: effective; F: field.  But simulated neurons in $X$ do not have Sensory Receptive Field (SRF) and Sensory Effective Field (SEF) because they only effect $Y$ and those in $Z$ do not have Motor Receptive Field (MRF) and Motoric Effective Field (MEF) because they only receive from $Y$.
 (b) The resulting self-wired architecture of DN with Occipital, Temporal, Parietal, and Frontal lobes.  
 Regulated by a general-purpose Developmental Program (DP), the DN self-wires by ``living'' in the physical world.  The $X$ and $Z$ areas are supervised by physics, including self, teachers, and other
 physical events.  Through the synaptic maintenance \cite{Wang11,Guo15}, some $Y$ neurons 
gradually lost their early connections (dashed lines) with $X$ ($Z$) areas and become ``later'' (early) $Y$ areas.  In the (later) Parietal and Temporal lobes, some neurons further gradually lost their connections with the (early) Occipital area and become rule-like neurons.  These self-wired connections give rise to a complex dynamic network, with shallow and deep connections instead of a deep cascade of areas.  
Object location and motion are non-declarative concepts and object type and language sequence
are declarative concepts \cite{Sun05}.  Concepts and rules are abstract with the desired specificities and invariances.  See Methods for why DN do not have any static Brodmann areas.  
}
 \label{FG:SWWnetFunctions3-2}
\end{figure*}

Next, we drop symbols altogether for our machine.  Why? A symbol is atomic, whose meanings are in the programmer's document, not told to the symbolic TM.  They are also too static for real-time tasks.
Suppose you, assisted by a symbolic TM, drive into a new country that uses a new language (e.g., new signs) but the programmer of your symbolic TM 
has not considered this new language.  Your biological brain immediately deals with the {\em patterns} (e.g., images) of new signs directly without the programmer's document because you can pull your car over and start to learn.  Namely, your brain starts to auto-reprogram itself.
But your symbolic TM in Fig.~\ref{FG:SWWnetFunctions3-2}(b) cannot because all its symbols are static and your programmer has left you!  Weng \cite{WengRepRev12} proved that your brain is free of symbols for a complexity reason. 

For auto-programming, we need a new theory that uses exclusively natural patterns (e.g., image patches of cars and signs).  The six necessary conditions are in the acronym GENISAMA below. 

{\bf GENISAMA TM:}  As illustrated in Fig.~\ref{FG:TM-GSM-GENISAMA}(c) it has a Developmental Network (DN) as its control and the real (physical) world as its ``tape''.   The DN has three areas, sensory $X$,  hidden $Y$ and motoric $Z$ with details shown in Fig.~\ref{FG:SWWnetFunctions3-2}.   We also use $X$, $Y$, $Z$ to denote the spaces, respectively, of the corresponding neuronal response patterns.   

If $X$ and $Z$ contain all sensors and effectors of an agent,  $Y$ models the entire hidden ``brain''.   
If $X$ and $Z$ correspond to a subpart of the brain areas, $Y$ models the brain area that connect $X$ and $Z$ as a two-way ``bridge''.   The computational meanings of the acronym GENISAMA are as follows:  

{\bf Grounded}: All patterns  $\z\in Z$ and $\x\in X$ are from the external environment (i.e., the body and the extra-body world), not from any symbolic tape.  

{\bf Emergent}: All patterns $\z\in Z$ and $\x\in X$ emerge from activities (e.g., images). All vectors $\y\in Y$ emerge automatically from $\z\in Z$ and $\x\in X$.   

{\bf Natural}:  All patterns $\z\in Z$ and $\x\in X$ are natural from real sensors and real effectors, without using any task-specific encoding, as illustrated in Fig.~\ref{FG:SWWnetFunctions3-2}.    

{\bf Incremental}:  The machine incrementally updates at times $t=1, 2, ... $.  Namely DN uses $(\z(t), \x(t))$ for update the network and discard it before  
taking the next $(\z({t+1}), \x({t+1}))$.  We avoid storing images for offline batch training (e.g., as in ImageNet) because the next image $\x({t+1})$ is unavailable without first generating and executing 
the agent action $\z(t)$ which typically alters the scene that determines $\x({t+1})$.  

{\bf Skulled:}  As the skull closes the brain to the environment, everything inside the $Y$ area (neurons
and connections) are initialized at $t=0$ and off limit to environment's direct manipulation after $t=0$.  

{\bf Attentive:} In every cluttered sensory image $\x\in X$ only the attended parts correspond to the current attended symbol set  $s$.  In every cluttered motoric image $\z \in Z$ only the attended parts correspond to the current state symbol $q$ (e.g., firing muscle neurons in the mouth and arms).  Two symbols correspond to a pattern (not necessarily connected, as in $s=\{\mbox{car2, pedestrian1\}}$). 

{\bf Motivated:}  Different neural transmitters have different effects to different neurons, e.g., resulting in (a)  avoiding pains, seeking pleasures and speeding up learning of important events and (b) uncertainty- and novelty-based neuronal connections (synaptic maintenance for auto-wiring) and behaviors (e.g., curiosity).

{\bf Abstractive:} Each learned concept (e.g., object type) in $Z$ are abstracted from concrete examples in $\z \in Z$ and $\x\in X$, invariant to other concepts  learned in $Z$ (e.g., location, scale, and orientation).  E.g., the type concept ``dog'' is invariant to ``location'' on the retina (dogs are dogs
regardless where they are).  Invariance is different from correlation: dog-type and dog-location are correlated (e.g., dogs are typically on ground).

{\bf The GENISAMA control as DN:} Assume a human knowledge base is representable by a grand TM,
whose FA control has alphabet $\Sigma = \{ \sigma_1, \sigma_2, ... , \sigma_n \}$, a set of states $Q=\{ q_1, q_2, ... , q_m\}$, and a static lookup table as its transition function 
$\delta: Q\times \Sigma \mapsto Q$.   The lookup table has $n$ columns for $n$
input symbols and $m$ rows for $m$ states.  Each transition of the FA control is from state $q_i$ and input $\sigma_j$, to the next state $q_k$, denoted as $(q_i, \sigma_j) \rightarrow q_k$, corresponding to the 
$q_k$ entry stored at row $i$ and column $j$, in the lookup table.   

Required by GENISAMA, let grounded $n$ (emergent) vectors $X = \{\x_1, \x_2, ... , \x_n\}$ represent the $n$ (static) symbols in $\Sigma$, so that $\x_j \equiv \sigma _j, j=1, 2, ... , n$ where $\equiv$ means ``corresponds to''.
Likewise, let $m$ (emergent) vectors $Z = \{\z_1, \z_2, ... , \z_m\}$ represent the $m$ (static) symbols in $Q$, so that $\z_i \equiv q _i, i=1, 2, ... , m$.  Thus, each symbolic transition (left, static) in FA corresponds to the vector mapping (right, emergent) in DN:
\[
[(q_i, \sigma_j) \rightarrow q_k] \equiv [(\z_i, \x_j) \rightarrow \z_k].  
\]
Because of the reasons in Weng \cite{WengRepRev12}, the lookup table for the human common-sense base is exponentially wide and exponentially high, but also extremely sparse. Yet, the right-side in the above equation uses only observed sparse entries emerged, where each entry corresponds to a neuron in DN. 

Denote $\dot \v = \v/\| \v \|$, i.e., normalizing the Euclidean length of $\v$. 

The neurons in $X$ and $Z$ are open to the environment, supervisable by the environment. 

Next, let 
the Grand TM in the environment teach the DN by supervising its $X$ and $Z$ ports while TM runs, one transition at a time in real time.  The DN has its brain area $Y$ area hidden (i.e., skulled). 

The simplest DN learns incrementally as follows.  Given each observation $(\z, \x)$ from the teacher 
TM, all $Y$ neurons compute their goodness of match. 
Each $Y$ neuron $(i,j)$ corresponds to an observed transition at the $(i,j)$ entry of the lookup table. 
In order to match both $\z$ and $\x$, it has a two-part weights $\v_{ij} =(\t_{ij},  \b_{ij})$. 
When the best match is not perfect explained below, $(\z, \x)$ is the left-side of a new 
transition; so DN incrementally adds one more $Y$ neuron by setting its $\t_{ij} = \z_i$ and $\b_{ij} = \x_j$.  
So, DN adds up to (finite) $mn$ hidden neurons, but typically much fewer because the lookup table is sparse.     

The top-down match value is $v_t = \dot{\t} \cdot \dot{\z}$; and bottom-up match $v_b = \dot{\b} \cdot \dot{\x}$.  We know that $\dot{\a} \cdot \dot {\b} = \cos \theta$, where $\theta$ is the angle between 
the two unit vectors $\dot{\a}$ and  $\dot{\b}$.  $\cos \theta =1$ is maximized if and only iff $\dot{\a} = \dot{\b}$, namely $\theta = 0$. The match between the current context input $(\z, \x)$ with the 
weight $(\t,  \b)$ of a  $Y$ neuron is the sum (or product) of the bottom-up and top-down match values, as its
pre-response value:
\[
f (\z, \x \;|\; \t, \b) =  v_t + v_b = \dot{\t} \cdot \dot{\z} + \dot{\b} \cdot \dot{\x} 
\]
Only the best matched $Y$ neuron fires (with response value 1), determined by a highly nonlinear competition:
\[
(i',j') = \arg \max_{(i, j) \in Y } f (\z, \x \;|\; \t_{ij}, \b_{ij}) = \arg \max_{(i, j) \in Y } \{ \dot{\t}_{ij} \cdot \dot{\z} +\dot{\b}_{ij} \cdot \dot{\x}\}. 
\]
All other loser $Y$ neurons do not fire (response value 0), because otherwise these neurons not only create more noise but also lose their own long-term memory (since all firing neurons must update using input).

The area $Z$ incrementally updates so that the firing $Y$ neuron $(i',j')$ is linked to all firing components (i.e., 1 not 0) in $\z_k$, so DN accomplishes every observed transition $(\z_i, \x_j) \rightarrow \z_k$, error-free, as proved in Weng \cite{WengIJIS15}.

Using the optimal Hebbian learning in Methods, Weng \cite{WengIJIS15} further proved that (1) the 
weight vector of each $Y$ neuron in the optimal (maximum likelihood) estimate of observed samples in 
$(X, Z)$, (2) the weight from each $Y$  neuron $(i', j')$ to each $Z$ neuron $k$ is the probability for $(i', j')$ to fire, conditioned on $k$ fired, and (3) overall, the response vectors $\y$ and $\z$ are both optimal (maximum likelihood).

Thus,  DN uses at most $mn$ $Y$ neurons, observes each symbolic transition $(q_i, \sigma_j) \rightarrow q_k$ in TM represented by vector transition $(\z_i, \x_j) \rightarrow \z_k$, and learns each error-free
if each input $(\z, \x)$ is noise-free.  If input $(\z, \x)$ is noisy, DN is optimal.   Namely, DN both 
``overfits'' and is optimal, regardless input is noisy or noise-free.  This is a new proof for TM  emerging from DN, shorter but less formal than Weng \cite{WengIJIS15}.

Attention corresponds to weights $\t$ and $\b$ partially connected with $Z$ area and $X$ area, respectively, --- thanks to naturally emerging patterns $\z$ and $\x$.

{\bf Auto-programming:} Consider two learning modes.  Mode 1:  Learn from a teacher TM supervised.   Mode 2: Learn from the real physical world without any explicit teacher.   
For early learning in Mode 1 to be useful for further learning in Mode 2,  assume that the patterns in Mode 1 are grounded in (i.e., consistent with) the physical world of Mode 2.

\begin{theorem}
\label{TM:Optimal}
By learning from any teacher TM (regular or universal) through patterns (Modes  1 and 2) with top-1 firing in $Y$, the DN control enables a learner GENISAMA TM to emerge inside it with the following properties.
\begin{enumerate}
\item Sufficient neurons situation: The GENISAMA TM is error-free for all learned TM transitions (Mode 1) and resubstitution of all observed physical experiences (Mode 2).   
\item Insufficient neurons situation: This happens when the finite $n$ $Y$ neurons have all been activated.    
The action at time $t+1$ is optimal in the sense of maximum likelihood (but not error-free) in representing the observed context space $(\z, \x)$, conditioned on the amount of computational resource $n$ and the experience of learning for all discrete times $0, 1, 2, ... , t$.  
\end{enumerate}
\end{theorem}

The proof sketch is available in Methods. 

Next, consider auto-programming for general purposes.  We represent each purpose as a TM.  Suppose a Grand Transition Table $G$ represents the FA control of a grand TM.   This $G$ contains a Universal TM $T_u$ and a finite  number of tasks as TMs, $T_i$, $i=1, 2, ... $.   Traditionally, $T_u$ is based on a (symbolic) computer language, but here $T_u$ can be in a (non-symbolic) natural language if it is GENISAMA.  

\begin{theorem}
\label{TM:auto-program}
A GENISAMA TM inside DN automatically programs for general purposes $T_i, i=1, 2, ... $, after it has learned a Universal TM $T_u$ and the related purposes $T_i, i=1, 2, ... $.  However, the DN algorithm (developmental program) itself is task-independent and language-independent (e.g., English or Chinese). 
\end{theorem}

The proof sketch is available in Methods. 

The Church-Turing thesis is a hypothesis that (A) a function on 
natural numbers is computable by a human using a pencil-and-paper method if and only if (B) it is computable by a TM.   This hypothesis is considered not provable because (A) involves a human 
who senses and acts in the real world whose process was not sufficiently formulated before to make a proof
possible.   The GENISAMA TM sufficiently formulates human auto-programming in the real-world and
therefore a full proof now becomes possible.  The following theorem extends the Church-Turing thesis.
\begin{theorem}
\label{TM:Church}
For any machine, natural or artificial, the following (a), (b) and (c) are equivalent in terms of computing engine: (a) Its GENISAMA augmentation learns and does auto-programing for general purposes in the real world.  (b) It computes all functions that are computable using a pencil-and-paper method.  (c) It computes all functions computable by TMs.
\end{theorem}

The proof sketch is in Methods. 
Therefore, it has been constructively proved (sketchy) in theory that a machine can perform auto-programming for general purposes.
These constructive proofs explain also how to.  The companion paper shows 
how to practically approach strong AI. 

\section*{Methods}

Before we discuss the detail of the new methods, let us first review major methods in the literature. 

In Artificial Intelligence, there are two schools, symbolic and connectionist \cite{Minsky91,WengRepRev12,Gome14}.   On one hand, the meanings of symbolic representations (e.g., Bayesian Nets \cite{Pearl86,Lake16}, Markov models \cite{Rabiner89,Puterman94}, graphic models by many) are static (before probability measures) in the human designer's mind and design documents, but are not told to the machine.  For example, such symbolic representations prevent the machine from learning the meanings of 
new concepts beyond those having already been statically handcrafted.  On the other hand, representations in artificial neural networks can emerge from activities but lack \cite{Minsky91,Gome14} clearly understandable logic, such as abstraction, invariance, and the hierarchy of relationships. 
For example, deep learning networks \cite{Fukushima75,WengCresIJCV97,Schmidhuber15,LeCun15,Mnih15,Fei-fei16} and other brain inspired network models \cite{Albus10,Eliasmith12} have shown impressive capabilities.  However, they lack a crucial function that appears to be necessary to scale up from a human fetus, to human infant, to human adulthood: a system automatically and directly learns from physical world for open domains.

The work there demonstrate, through a constructive (sketchy) proof, that this requires a framework that is beyond {\em from-pixels-to-handcrafted-text} \cite{Fukushima75,WengCresIJCV97,Albus10,Eliasmith12,Schmidhuber15,LeCun15,Mnih15,Fei-fei16}, 
that is not an emergent-and-symbolic-hybrid \cite{Albus10,Mnih15,Fei-fei16,Sun07} either, but rather whose representations are {\em exclusively emergent and there are no symbols in DN at all}.  In such a drastically different architecture and its representations, detection, recognizing, {\em attention}, and action on individual object(s) all take place in parallel in cluttered environments where background pixels are often many more than object pixels.  

Unique in this regard, representations in DN \cite{WengWhy11}, supported by a series of embodiments called Where-What Networks, WWN-1 through WWN-9, take the best of both AI schools: not only emergent, but also logic; not only logic, but also complete in the sense of TM \cite{WengIJIS15}.   However, it is unknown whether a DN can auto-program for general purposes.  The work here takes up this fundamental issue: Can machines automatically learn to think for general purposes --- not relying on any handcrafted symbols, let alone any world models?

In Natural Intelligence, the task-nonspecificity of an innate Developmental Program (DP) is highly debatable \cite{Chomsky78,Elman93,Elman97,QuartzSejnowski97,Bates98,Pinker09,WengScience}.  
It is still unclear computationally how a DP can regulates a neural network, natural or artificial, to enable the network to auto-program concepts and rules from the cluttered physical environment.   This paper
proposes a computational theory for that without claiming to be biologically complete. 
Since spontaneous neuronal activities are present prenatally \cite{Katz96,Feller96},
the activity-dependent wiring mechanisms here could take place both before and after birth.  The model proposed here contains supports for both nativism and empiricism, but more explicit and precise computationally. 

Theories of machine-learning {\em logical} computations have been fruitfully studied (e.g., \cite{Valiant84,Valiant00,Jordan15}) to deal with {\em propositions} and {\em predicates}, whose answers are, yes or no (e.g., fraud or not fraud), but not both.   The {\em full automation of machine learning} in the
real world  
--- e.g., the emergence of representations (i.e., skull closed, through lifetime learning of an open series of unpredictable tasks) and automatic scaffolding (i.e., early-learned simple skills 
automatically assist later learning of more complex skills) ---
has not received sufficient attention.  However, this is the way for brains to autonomously learn from infancy to adulthood.  Since the Autonomous Mental Development (AMD) direction was 
proposed in \cite{WengScience}, a major progress in this direction, represented by the Where-What Networks
\cite{WengWhy11} as embodiments of DN, has not received sufficient attention either (see, e.g., recent reviews \cite{LeCun15,Schmidhuber15,Jordan15}).  However, the full automation of machine learning seems to be a practical way for machines to become as versatile as a 3-year-old human child in the three well acknowledged
bottleneck areas of AI --- vision, audition and natural language understanding.  The theory here 
is necessary for the {\em full automation of machine learning}. 

Through Finite Automata \cite{Hopcroft06,Martin11}, we extend such logic to spatiotemporal 
sensorimotor {\em actions}, to deal with kinds of intelligence that require behaviors, 
including vision, audition (recognition of not only speech, but also music, etc.), natural language acquisition,  and vision-guided navigation.  Namely, not only logic, but also interactive actions where logic is a special case. 
This letter focuses on theory and algorithm; a companion paper submitted
concurrently focuses on experiments.  We will see that auto-programming is necessary 
for the full automation of machine learning and also explains how a human adult can do so.  

There are two types of approaches to modeling a brain.  The first type assumes that the genome rigidly  dictates all Brodmann areas (e.g., V1 and V2) inside the brain so such a model starts with static existence of the Brodmann areas such as those reported by \cite{FellemanVanEssen91}.  The second type, which this model belongs to, does not assume so.   Such a model does not have static brain areas because the formation of, and the existence of, brain areas depend on activities, as demonstrated by the
following studies:  
\begin{enumerate}
\item Cells in the V1 area selectively respond to the left eye, the right eye, and both eyes in a
normal kitten; but they respond only to one eye if the other 
eye is closed from birth \cite{WieselHubel65}.  Namely, where an area connects from is plastic. 
\item A pathway amputation early in life enabled the {\em auditory} cortex to receive {\em visual} signals through actively growing neurons so that the {\em auditory} cortex emerged {\em visual} representations and the animal demonstrated
{\em visual} capabilities using rewired {\em auditory} cortex \cite{VonMelchner00}.  Namely, what an area does is plastic. 
\item The visual cortex is reassigned to audition and touch in the born blind \cite{Voss13}.  Namely, visual areas may completely disappear. 
\end{enumerate}
Therefore, the DP of this model does not specify Brodmann areas.  It enables ``general purpose'' neurons to wire, trim, and re-wire.  But the experimental demonstration for the formation of the spinal cord and the detailed Brodmann areas in the DN, as well as the plasticity thereof, remains to be future work.   
 
The following questions are:  Is there a central controller in the brain? Is there a Master Map in the brain?
Does the genome rigidly specify features or instead features emerge from both prenatal (i.e., innate) and 
postnatal development?   Does the brain uses convolution --- replication of neural weights across different neurons?    Does the brain consist of a rigid deep cascade of processing modules?    Inspired by the above plasticity studies, the new theory here does not assume the static existence of a {\em Master Map}  proposed by Anne Treisman 1980 \cite{Treisman80,Treisman86} and used by 
others \cite{Anderson87,Olshausen93,Tsotsos95,Itti98}.  Such a Master Map requires a central controller who is already intelligent, so that it selects every 
attended object image-patch from each figure-ground-mixed image on the retina and feeds only attended figure patch into the Master Map.  In the Master Map, the location and scale of the figure are normalized
so that remaining issue is only classification.    In some sense, the ``normalization'' that we hope 
is performed by the motoric area $Z$ in the theory area, but $Z$ is not a feature map but an action map. 

In the Computational Vision literature, such a central controller is a human.   Bottom-up features
\cite{Lee99,WengLCA09} and their saliencies have been proposed \cite{Olshausen93,Itti98,IttiKoch01} to partially serve the role of this central controller --- a salient patch is fed into the Master Map.  Another example of human central controller is Cresceptron  \cite{WengIJCNN92,WengCresIJCV97} and much later work where the human trainer manually draws a polygon on the sensed image that segments a human attended figure from the ground so that the system learns bottom-up from only pixels inside the polygon.   
However, the brain anatomy \cite{FellemanVanEssen91,Kandel12} 
appears to allude to us that the brain network contains various shallow and deep circuits in which bi-directional connections are almost everywhere, not just a cascade.   The new theory here assumes that neurons automatically connect, not only bottom-up, but also top-down \cite{Li04,Buschman07,Reddy07,Saalmann07,Luciw10,WengIEEE-IS2014} and lateral, all using accumulated statistics in neuronal activities.     

Deep learning convolutional networks \cite{Fukushima83} with max-pooling \cite{WengCresIJCV97,Serre07} and other techniques \cite{LeCun98,Krizhevsky12,SchmidhuberNNreview14,LeCun15,Mnih15,Schmidhuber15,Jordan15} have shown their power in pattern classification --- output class labels. They all imposed a cascade of processing modules/layers.   The max-pooling is meant to reduce 
the location resolution from each early layer to the next layer to avoid the exponential explosion of the template size of convolution.  However, the amount of computation can be contained by enabling each $Y$ feature neuron --- in early and later layers --- to have two 
input sources, bottom-up from $X$ and top-down from $Z$.  Therefore, not only location ``resolution'' is automatically reduced from early to later $Y$ areas 
through the ventral pathway (for outputting class information), but also the type ``resolution'' is also 
automatically reduced from early to later $Y$ areas through the dorsal pathway (for outputting location- and manipulation-information).    This top-down and bottom-up two-input $(\z, \x)$
architecture seems to provide a more flexible architecture for dealing with pattern recognition,
either from monolithic or from cluttered scenes.    Such a non-cascade network seems to be consistent with  
neuroanatomical studies reviewed in \cite{FellemanVanEssen91}.   

{\bf Agent FA} To understand how a symbolic state can abstract both spatial and temporal contexts, consider Task 1:  Produce the truth-value of  an input logic-AND expression like:
\[
T \wedge F \wedge T \wedge T
\]
written on a tape. 
A regular FA only has an input string not a tape but this tape-view is useful next.  We allow the tape head to read the input sequence by moving only right, a symbol at a time, and to read only. In general, such a logic-AND expression consists of a finite number of input symbols from alphabet $\Sigma = \{ T, F, \wedge \}$, where $T$ and $F$ represent
true and false, respectively, and $\wedge$ denotes logic AND.   Let $Q$ be the set of states of the FA handcrafted by a human programmer. 

The control of an Agent FA is a function $\delta: Q \times \Sigma \mapsto Q$, a mapping from domain
$Q \times \Sigma$ to codomain $Q$.  

Table~\ref{TB:ANDSymbol} gives the control for Task 1, where the meaning of each state is denoted by the subscript of $q$.  The patterns for $\x$ and $\z$ will be needed later in the paper.   At row $q$ and column $\sigma$ is the next state $q'= \delta(q, \sigma)$, or denoted
graphically as $(q, \sigma) \rightarrow q'$. 
E.g., at the initial state $q_0$, receiving
an input $T$, the next state is $q_T$ to memorize the context $T$.  This gives $\delta (q_0, T) = q_T$.
Similarly,  $\delta (q_T, \wedge) = q_{T\wedge}$ to memorize context $T\wedge$. Then,
$q'=\delta(q_{T\wedge},F)=q_F$, because $T\wedge F=F$. 
The transition sequence for the above input $T \wedge F \wedge T \wedge T$ is
\begin{equation}
q_0 \stackrel{T}{\longrightarrow} q_T \stackrel{\wedge}{\longrightarrow}  q_{T\wedge}  \stackrel{F}{\longrightarrow}  q_{F} 
\stackrel{\wedge}{\longrightarrow}  q_{F\wedge}  \stackrel{T}{\longrightarrow} q_F  \stackrel{\wedge}{\longrightarrow}  q_{F\wedge}  \stackrel{T}{\longrightarrow} q_F .
\label{TB:task1trans}
\end{equation}
The state $q_-$ represents that the input sequence is an invalid logic-AND expression,  e.g., $\wedge TF$ or $ F\wedge \wedge$.

\begin{table}[tb]
\caption{Control $\delta$ of FA for Task 1 and the pattern representations of state $q$ and input $\sigma$.}
\begin{center}
\begin{tabular}{|cc|lccc|}
\hline
\multicolumn{2}{|c|}{$\delta(q,\sigma)$}  & Input $\sigma$ &  $T $ 		&  $F$ 		&$ \wedge$   \\
 \cline{1-2}
State $q$ & State pattern $\z$                      & Input pattern $\x$ & 010                    & 011              & 100   \\
 \hline
$q_0 $ & 001	& & $q_{T} $	& $q_{F}$   	&    $q_{-}$    \\
$q_{T}$ & 010	& &  $q_{-}$    		& $q_{-}$    		& $ q_{T\wedge} $ \\
$q_{F}$ &  011	&&  $q_{-}$    		& $q_{-}$   		& $ q_{F\wedge} $ \\
$ q_{T\wedge}$ & 100 &  & $q_{T}$ & $q_{F}$ &  $q_{-}$ \\
$ q_{F\wedge}$ & 101 & &  $q_{F}$ & $q_{F}$ & $q_{-}$  \\
$q_{-}$ & 110 	& & $q_{-}$ 				&$q_{-}$ 			&	$q_{-}$ \\
\hline
\end{tabular}
\end{center}
\label{TB:ANDSymbol}
\end{table}%

This is  {\em temporal} abstraction from examples:  Each state memorizes only the necessary context information for the specific Task 1.  The abstraction in the previous state facilitates the abstraction of the next state.
In natural language acquisition, the temporal context for each state is similar but more complex. 
 
As {\em spatial} abstraction from examples, we can extend the Task 1 so as to handle symbol $\tau$ as $T$ and $\phi$ as $F$, respectively.   All we need to do is expand Table~\ref{TB:ANDSymbol} by adding two additional columns for inputs $\tau$ and $\phi$, respectively, but using the same next states as $T$ and $F$.  During
vision-guided autonomous driving, different traffic semaphores are like $T$ and $\tau$ here, but more complicated. 

Thus, both spatial and temporal abstractions take place concurrently in each transition: $(q, \sigma) \rightarrow q'$.  We will see in Table~\ref{TB:2TasksPattern} below that when a brain applies this mechanism to patterns,  the brain deals with space and time in a unified way, independent of meanings.  

It is useful below to see how the control implements Table~\ref{TB:ANDSymbol}:  Given any state $q$ and input $\sigma$, the control finds the matched row $q$ at row and matched column $\sigma$.  The table cell stores the information for the next state $q'$.  Below, each table cell will correspond to a neuron whose inputs are the original patterns (not symbols) of $q$ and $\sigma$ as shown in Table~\ref{TB:ANDSymbol}. 

In Task 1, inputs $T$ and $T\wedge T$ lead to the same state $q_T$. 
This process requires state design and equivalent-state finding for spatiotemporal abstraction.
Handcrafted by humans, such {\em symbolic representations} are logic and clean \cite{Minsky91,Gome14}.
But they become manually intractable and thus error-prone (brittle) when the transition table has  exponentially many rows and columns for natural languages or autonomous driving  \cite{Lenat95,Saygin00,WengRepRev12,Harnad14,You15}.  Below, we will see that the natural world can supervise each transition $(q, \sigma) \rightarrow q'$ but using directly patterns which are without human handcrafting.  

{\bf Attentive Agent FA}  An Attentive Agent FA has a set of input symbols, called alphabet $\Sigma$, of a finite size.   At each time $t$, $t=1, 2, ... $, it attends to a set $\Sigma(t)\subset \Sigma$ of symbols from the {\em symbolic} environment $E(t)$ \cite{Posner80,Koch85,Desimone95,Corbetta02,IttiReesTsotsos,Ji10,Luciw10}.   The set $\Sigma(t)$ can be a 2-D patch of text (e.g., of this page) or a substring of the input sequence (e.g.,  $ T \wedge F $ of $T\wedge T \wedge F \wedge T$).    The state/action from the machine may change the environment and also the next sensed $\Sigma(t+1)$.  

In Task 1, the single-letter-right-only scan is only one of many ways of the Attentive Agent FA.   
For an unconstrained Attentive Agent FA, a human programmer must handcraft a large lookup table $\delta: Q \times 2^\Sigma \mapsto Q$ so that the output state $q(t)\in Q$ at every time $t$ enables the Attentive Agent FA to sequentially complete the given task.  The number of columns of 
the transition table of $\delta$ is exponential in the size of $\Sigma$ because of the power set $2^\Sigma$ in  the domain of $\delta$.  The number of rows, the size of $Q$, may potentially also be exponential in the size of $\Sigma$.

Task 1 does not need this freedom of attention.  However,  an Attentive Agent FA is useful
for the more challenging Task 2:  Produce the truth value of an input sequence that includes logic operators $\wedge$, $\vee$, and
parentheses, such as:
\[
T\wedge ((T \vee F) \wedge T \vee F ).
\]

It is known \cite{Hopcroft06,Martin11} that the single-letter-right-only scan can still accomplish Task 2 if the machine has an infinite-size stack so that it can store an unbounded number of left parentheses.  

If the machine can write onto the tape, the machine is a TM illustrated in Fig.~\ref{FG:TM-GSM-GENISAMA}(a) without the need for the stack.   A human can program a TM to perform Task 2 \cite{Hopcroft06,Martin11}.  

{\bf Proof of Theorem~\ref{TM:AttentiveFA}:}   The control of a TM has a transition function $\delta: Q\times \Gamma \mapsto Q \times \Gamma \times D$, where $Q$, $\Gamma$ and $D=\{R, L, S\}$ are the sets of states, the tape alphabet, and head moves, respectively.  We extend $\delta$ to $\delta': Q'\times \Gamma \mapsto Q'$ which is the form of  the control of an Agent FA, where
$Q' = Q \times \Gamma \times D$.  We have proved that the control of a TM is an Agent FA.
The above extension of domain 
from $Q\times \Gamma$ of $\delta$ to $Q'\times \Gamma$ of $\delta'$ means that for all $q' = (q, \gamma, d)\in  Q \times \Gamma \times D$ and $\gamma' \in \Gamma$, $\delta ' (q',  \gamma' ) = \delta (q, \gamma')$.  Namely $\delta'$ is independent of, or does not attend to, the last written symbol $\gamma$ and head move $d$ in its domain (as they are often encoded in state).   But this attention is dynamic, as the head can 
scan multiple positions to reach a state.  Namely, the control is an Attentive Agent FA.  This ends the proof.

A {\bf Grounded Symbolic Machine} illustrated in Fig.~\ref{FG:TM-GSM-GENISAMA}(b) can deal with additionally input patterns (image, lidar, sound, etc.), but it cannot automatically program for general purposes because it still requires a human programmer to handcraft the meanings of every input symbol $\sigma$ used to represent its input features, internal states $q$, and output actions.   A probability 
version \cite{Russell03} alleviates the uncertainty in such symbols, but cannot address the
inadequacy of static symbols to represent a new town, or a new situation (e.g., rain or hacker laser 
\cite{Harris15} for lidar) 

Neural networks (e.g., \cite{Rumelhart86,McClelland86} and many others) have been using patterns directly; but traditional neural networks do not have grounded symbol-like capabilities \cite{Harnad90,Minsky91,Sun05}.  Bridging this gap requires a machine to learn not only symbol-like concepts directly from non-symbols but also attention rules --- to quickly capture relevant patches (e.g., $s$ in Fig~\ref{FG:TM-GSM-GENISAMA}(b)) that are necessary for immediate action and disregard remainders (e.g., $\bar{s}$ in Fig~\ref{FG:TM-GSM-GENISAMA}(b)).  
Such attention rules are implicit; we often attend without knowing reasons.   The intractability of handcrafting such implicit rules demands general-purpose auto-programming. 

{\bf GENISAMA TM.}  Early neural network models for FA \cite{Frasconi95,Frasconi96,Omlin96} and for TM \cite{Siegelmann95,Siegelmann95b} are laudable for computing the automata mapping  
using networks but they used special encodings and do not learn, having none of GENISAMA.  
E.g., the TM in \cite{Siegelmann95,Siegelmann95b} used 2-D registered inputs (one signal line and the 
other line means the presence of signal in the signal line).  
In contrast, the inputs in $X$ here are unregistered
(e.g., an object can appear anywhere in the image) and cluttered (typically more noise/background dimensions than signal dimensions).   The TM in \cite{Siegelmann95b} extends to irrational numbers 
using infinitely long numbers, but words of a finite length should be sufficient 
for a practical GENISAMA TM (e.g., it recognizes and understand the irrational number $\sqrt{2}$ by the shape $\sqrt{2}$ and its rules instead of the infinitely long number).

The environment of the control DN is divided into internal environment (e.g., the network that learns an equivalent lookup table for the control but more efficient than exponential) and external environment.  The external environment includes the body of the agent and extra-body environment.  

Each area of DN control may have multiple subareas:  $X$ may contain two retinae, two cochlear hair cell arrays, somatosensory arrays, and receptor arrays of other sensory modalities.  $Z$ may contain muscle arrays for the mouth, the arms, and effectors of other motor modalities.   $Y$ as the internal representation of the control senses
the pattern in its input space $Z\times X =\{ (\z, \x) \;|\; \z \in Z, \x \in X\}$ to produce $Y$ patterns.  In turn, 
each of $X$ and $Z$ uses the $Y$ pattern to further predict the pattern in themselves.   Motivated by brain plasticity discussed below, we let subareas in $Y$  to automatically emerge (like Brodmann areas \cite{Kandel12,Gluck13}) instead of statically handcrafted.   

Suppose, in Table~\ref{TB:flow}, a GENISAMA TM learns from a teacher TM.  
The teacher is via its Attentive Agent FA control and the learner is via its DN control.
For clarity, suppose that each area of DN finishes an update computation in a unit time.  Then, let every area of DN run in discrete times $t=0, 1, 2, ... $ in parallel.    The Attentive Agent FA does not have any internal area because it is symbolic---a lookup table is sufficient.  The DN takes an additional unit time for the $Y$ area to update and interpolate.  That is why Table~\ref{TB:flow} only needs to specify the Attentive Agent FA at even time instants. 

\begin{table}[htb]
\caption{The correspondence between the symbolic Attentive Agent FA and the DN control of GENISAMA TM}
\begin{center}
\begin{tabular}{ccccccccccc}
$ \left[
\begin{array}{c}
  q(0)   \\
  \sigma(0)  
\end{array}
\right] $
&
$ \rightarrow $
&
$ \rightarrow $
&
$ \rightarrow $
&
$ \left[
\begin{array}{cc}
   \emptyset & \overline{q(2)}  \\
  \emptyset & \underline{\sigma(2)} 
\end{array}
\right] $
&
$ \rightarrow $
&
$ \rightarrow $
&
$ \rightarrow $
&
$ \left[
\begin{array}{cc}
   \emptyset & \overline{q(4)}  \\
   \emptyset & \underline{\sigma(4)}
\end{array} 
\right] $
&
$ \rightarrow $
&
$ \cdots $
\\
\\
$\left[
\begin{array}{c}
  \z(0)   \\
  \x(0)   
\end{array}
\right]$
&
$\rightarrow$
&
$\y(1)$
&
$\rightarrow$
&
$\left[
\begin{array}{cc}
  \emptyset &  \overline{\z'(2)}  \\
  \emptyset &  \underline{\x'(2)}
\end{array}
\right]$
&
$\rightarrow$
&
$\y(3)$
&
$\rightarrow$
&
$\left[
\begin{array}{cc}
  \emptyset  &  \overline{\z'(4)}  \\
  \emptyset  &  \underline{\x'(4)}
\end{array}
\right]$
&
$\rightarrow$
&
$ \cdots $
\\
\\
$\y(0)$
&
$\rightarrow$
&
$\left[
\begin{array}{cc}
  \emptyset & \overline{\z'(1)}  \\
  \emptyset & \underline{\x'(1)}
\end{array}
\right]$
&
$\rightarrow$
&
$\y(2)$
&
$\rightarrow$
&
$\left[
\begin{array}{cc}
  \emptyset & \overline{\z'(3)} \\
  \emptyset & \underline{\x'(3)} 
\end{array}
\right]$
&
$\rightarrow$
&
$\y(4)$
&
$\rightarrow$
&
$\cdots  $
\\
\end{tabular}
\end{center}
\label{TB:flow}
\end{table}

In Table~\ref{TB:flow}, the first row is the time flow of the Attentive Agent FA control of the teacher TM where 
$q(t)$ and $\sigma(t)$, $t=0, 2, 4, ... $, are the {\em attended} state/action and input, respectively.  

A traditional FA does not predict input at all (see Eq.~\eqref{TB:task1trans}), but we require an FA or Attentive Agent FA to predict not only the next state $q'$ but also the next input $\sigma'$.   New here is that we use $q$ and $\sigma$ both to predict also $\sigma'$: 
\begin{equation}
\left[
\begin{array}{c}
  q  \\
  \sigma   
\end{array}
\right]
\rightarrow
\left[
\begin{array}{c}
  q'  \\
  \sigma '
\end{array}
\right].
\label{EQ:FAtransition}
\end{equation}
The more obstructively complete $q$ is, the better the prediction for $\sigma'$. 
When such a prediction of $\sigma'$ is not unique, the agent is {\em immature} to 
explain the attended environment.   A calf might be mature in terms of finding food, but immature in terms of avoiding its predators. 

With the real time indices in Table~\ref{TB:flow}, the framework of Bayesian Networks \cite{Pearl86,Russell10} can be applied to the Attentive Agent FA while avoid 
cyclic graphs in graphical models because each cycle occurs at a different time instance above.  This deals with the problem of cyclic graphs that static graphic models have avoided.

In each $2\times 2$ array inside Table~\ref{TB:flow}, the first column is predicted by the control and the second has been affected by supervision from the 
environment.  $\emptyset$ means an empty set---the prediction is undetermined.  An underline
for input (e.g., $\underline{\sigma(2)} $) or overline for state/action (e.g., $\overline{q(2)} $) means the environment supervises and the supervision is different from what the control predicted.  

The control DN of GENISAMA TM learns from the teacher TM by taking one taught 
$(q', \sigma')$ at a time in Eq.~\eqref{EQ:FAtransition}, but dealing with patterns directly.  Symbolically, it learns a mapping $\delta: Q\times \Sigma \mapsto Q\times \Sigma$ from the teacher TM (e.g., mother or school instructor).

The DN uses original patterns $\z(t)$ and $\x(t)$ whose {\em attended parts} correspond to symbols $q(t)$ and $\sigma (t)$, respectively, denoted as $a(\z(t)) \equiv q(t) $ and $a(\x(t)) \equiv \sigma (t)$ where 
function $a$ is a dynamically learned function that marks off the unattended components. 

Running at times $t=0, 1, 2, 3, 4, ... $, the 2nd and 3rd rows in Table~\ref{TB:flow} are two flows 
that run in parallel to predict the corresponding patterns $\x, \y, \z$ in all the three areas of the DN. 
The $Y$ area takes input from $(\z, \x)$ to produce a response vector $\y$ which is then used by $Z$ and $X$ areas to predict $\z$ and $\x$ respectively:
\begin{equation}
\left[
\begin{array}{c}
  \z  \\
  \x   
\end{array}
\right]
\rightarrow
\y
\rightarrow
\left[
\begin{array}{c}
  \z'  \\
  \x' 
\end{array}
\right]
\label{EQ:DNtransition}
\end{equation}
where the first $\rightarrow$ denotes the update in the left side using the left side 
as input.  Like the FA, each prediction in  Eq.~\eqref{EQ:DNtransition} is called a {\em transition}.  
The same principle is also used to predict the binary (or real-valued) $\x'\in X$ in Eq.~\eqref{EQ:DNtransition}.   The quality of prediction 
depends on how state/action $\z$ abstracts the external world sensed by $\x'$. 

{\bf Learning}  As the simplest version, we use a highly recurrent, winner-take-all  
computation to simulate parallel lateral inhibition in $Y$: the $Y$ area with $n$ neurons 
responds with $\y= (y_1, y_2, ... y_n)$ where 
\begin{equation}
y_j = \begin{cases} 1 & \mbox{if $j=\underset{1\le i \le n}{\operatorname{argmax}} 
\{ f_i( \z, \x )\}$}\\
0 & \mbox{otherwise}
\end{cases}
\label{EQ:Ycompute}
\end{equation}
$j=1, 2, ... n$, where each $f_i$ measures the goodness of match between
its input patch in $(\z, \x)$ and its weight vector $(\z_i, \x_i)$.  
The Hebbian learning together with the
synaptic maintenance explained below initialize, update, cut-and-grow, all weight vectors $(\z_i, \x_i)$, $i=1, 2, ..., n$,  resulting the rich connections illustrated in Fig.~\ref{FG:SWWnetFunctions3-2}.

Let the FA control of the teacher TM have $r$ rows and $c$ columns.   Suppose that in the
learner GENISAMA TM,  the $Y$ area has at least $n=rc$ $Y$ neurons.  For Table~\ref{TB:ANDSymbol}, $n=rc=6\times 3=18$.   
Without any central controller, all $Y$ neurons start with random weights at time $t=0$.   
At each time $t$, $t=1, 2, ... $, only the winner $Y$ neuron fires at response value 1 and incrementally 
updates its weight vector $(\z_i, \x_i)$ as the vector average of attended part of $(\z, \x)$.  Then, the $i$-th $Y$ neuron memorizes perfectly the $i$-th distinct input pair $(\z, \x)$ observed in life.   

The learner GENISAMA TM is taught by the teacher TM through the supervision of $(\z, \x)$.  Each $Z$ neuron represents a unique component in $\z$.   
It should fire at 1 (instead of not firing at 0) if and only if it 
has been taught to fire right after the firing $Y$ neuron.  This is true regardless in how many $Z$ patterns this $Z$ neuron appears.   In general, if the number of $Y$ neurons is insufficient and
input in $(\z, \x)$ has noise,  the weight from the firing $Y$ neuron to each $Z$ neuron is the incrementally updated probability for the pre-synaptic $Y$ neuron to fire conditioned on that the post-synaptic $Z$ neuron fires. 

Therefore, the roles of working 
memory and long-term memory in each area $Y$ are dynamic --- the firing neurons are the current working memory and all other currently non-firing neurons  are the current long-term memory.  In this way, it is always the best matched neurons to update while other non-firing neurons keep their memory intact.  When the number of $Y$ neurons is large, the finite-size DN appears to never run out of memory because the
top-matched neurons are near and the forgetting/update is for the nearest memory only.

{\bf Top-k and brain areas}  In general, top-$k$ $(k>1)$ neurons fire, as members of a distributed $Y$ ``committee'' in which only 
$k$ experts fire to vote, as illustrated in Fig. \ref{FG:SWWnetFunctions3-2}(b).  The top-$k$ mechanism itself is not biologically plausible, but it simulates mutual inhibitions among neurons (see
Fig.~\ref{FG:SWWnetFunctions3-2}(a)) so that much fewer $Y$ neurons fire in the presence of mutual inhibitions (see sparse coding idea \cite{Olshausen96}).  

The top-$k$ voting in Eq.~\eqref{EQ:DNtransition} was called the ``bridged-islands'' model \cite{WengIJCNN14}.  In general, the ``bridge'' $Y$ area can be considered any brain area where 
neurons fire to be used by its connected islands --- top islands $Z$ and bottom islands $X$.  The complex brain network hierarchy (e.g., see \cite{FellemanVanEssen91}) is not a cascade as modeled in deep learning.   Each area is a bridge that provides feature detectors for all neurons that are statistically correlated through excitatory connections and anti-correlated through inhibitory connections (see \cite{WengNAI12}).
Consequently, the sensory end of DN is the most concrete, having 100\% sensory content and 0\% motoric content.   The motor end of DN is the most abstract, 0\% sensory and 100\% motoric.   Other areas in DN are in-between, developing  intermediate abstract features that correspond to intermediate invariances (see, e.g., Figs. 6.5 and 6.6 in \cite{WengNAI12}).  This avoids the forced feedback loop in electrical engineering --- from the most abstract $Z$ back to the most concrete $X$ (see, e.g., \cite{Mnih15}).

{\bf Dynamic learning modes} The open ports $Z$ and $X$ are supervised or
free, depending on the external and internal environments.  By ``supervised'', we mean that, as soon as the port predicts a pattern as the left-column in each $2\times 2$ array in Table~\ref{TB:flow}, the external world overrides it as the right column of the $2\times 2$ array.  Otherwise, the $Z$ port is ``free'', predicting/generating actions from within.  If the ``eye'' is closed, the $X$ port is not supervised by the external 
environment and the $X$ port predicts
``mental images''.  (But ``mental images'' can, in principle, emerge also from early subareas in $Y$ like LGN, V1 etc. or written on a piece of paper through actions in $Z$, not requiring the eyes to close.)

Never directly supervised, the closed $Y$ uses {\em unsupervised learning}---optimal Hebbian learning explained below, although the agent's action maybe supervised by a teacher through the $Z$ area.  Namely, the body of the agent always supervises DN, but the $Y$ area always uses unsupervised learning!

Let us look at the example in Eq.~\eqref{TB:task1trans}.  We must let $q$ and $\sigma$ predict in parallel:
\begin{eqnarray}
\left[
\begin{array}{c}
  q_0   \\
 \underline{T} 
\end{array}
\right] 
& \rightarrow &
\left[
\begin{array}{cc}
  \emptyset & \overline{q_T}  \\
  \emptyset & \underline{\wedge} 
\end{array}
\right] 
\rightarrow
\left[
\begin{array}{cc}
  \emptyset & \overline{q_{T\wedge}}  \\
  \emptyset & \underline{F} 
\end{array}
\right] 
\rightarrow
\left[
\begin{array}{cc}
  \emptyset & \overline{q_{F}}  \\
  \emptyset & \underline{\wedge} 
\end{array}
\right] 
\rightarrow
\left[
\begin{array}{cc}
   \emptyset & \overline{q_{F\wedge}}  \\
  \emptyset & \underline{T} 
\end{array}
\right] \nonumber \\
& \rightarrow &
\left[
\begin{array}{cc}
  \emptyset & \overline{ q_{F}}  \\
  \emptyset & \underline{\wedge} 
\end{array}
\right] 
\rightarrow
\left[
\begin{array}{cc}
  q_{F\wedge} & q_{F\wedge}  \\
  T & T 
\end{array}
\right] 
\rightarrow
\left[
\begin{array}{c}
  q_{F}  \\
 \wedge 
\end{array}
\right] .
\label{EQ:symbolTransitions}
\end{eqnarray} 
where the last two predictions are perfect because of two reasons:  (a) the two predictions of state are unique due to the teacher consistency; (b) the two predictions of input are
unique since the learner is still naive ---  only 
$T$ follows $(q_F, \wedge)$ and it has not seen illegal input.  (Without better states that model the physical causality of input sequence, 
such symbolic prediction of input is not guaranteed to be unique.)

Using the patterns in Table~\ref{TB:ANDSymbol}, which are meaningless here but should correspond
to naturally emerging images, the DN learns the above teacher sequence one transition at a time, but through original patterns only and via $Y$ neurons:
\begin{eqnarray}
\left[
\begin{array}{c}
 001   \\
  \underline{010}  
\end{array}
\right] 
& \rightarrow y_1 \rightarrow &
\left[
\begin{array}{cc}
  \emptyset & \overline{010}  \\
  \emptyset & \underline{100} 
\end{array}
\right]
\rightarrow y_2 \rightarrow 
\left[
\begin{array}{cc}
  \emptyset & \overline{100}  \\
  \emptyset & \underline{011} 
\end{array}
\right] 
\rightarrow y_3 \rightarrow 
\left[
\begin{array}{cc}
  \emptyset & \overline{011}  \\
  \emptyset & \underline{100} 
\end{array}
\right] 
\rightarrow y_4 \rightarrow
\left[
\begin{array}{cc}
  \emptyset & \overline{101}  \\
  \emptyset & \underline{010} 
\end{array}
\right]  \nonumber \\
& \rightarrow y_5 \rightarrow &
\left[
\begin{array}{cc}
  \emptyset & \overline{011}  \\
  \emptyset & \underline{100} 
\end{array}
\right] 
\rightarrow y_4 \rightarrow
\left[
\begin{array}{cc}
  101 & 101  \\
  010 & 010
\end{array}
\right] 
\rightarrow y_5 \rightarrow
\left[
\begin{array}{c}
 011  \\
 {100} 
\end{array}
\right] 
\label{EQ:patternTransitions}
\end{eqnarray}
where $y_i$, $i=1, 2, ..., 5$, corresponds to the first five initialized $Y$ neurons.  
Two $Y$ neurons $y_4$ and $y_5$
predict perfectly when the same (or similar) pattern of $(q, \sigma)$ appears again.   
The term
``similar'' means interpolation that is impossible in Eq.~\eqref{EQ:symbolTransitions}.  

{\bf Attention} For simplicity, we have assumed above that $\z$ and $\x$ do not contain unattended parts. 
Of course,  in general the prediction of $\x$ pattern can cover fewer than all sensory
bits (3 bits above), amounting to experienced-based global-or-local sensory attention --- 
predicted bits are attended.  Namely, how the learner machine attended in the past ``lifetime'' shapes how it likely attends in the future ``lifetime''.

This example shows the model's separation of DP mechanisms (i.e., table lookup) from the meanings of the learned task.  Namely, the human programmer of the DP does not need to know the
meanings of patterns in Eq.~\eqref{EQ:patternTransitions} that emerge.  A regular TM and 
a Universal TM differ in the meanings of input symbols and state symbols, but they use the same domain $Q'\times \Sigma$ and codomain $Q'$ (i.e., table lookup) for their control function $\delta'$. Therefore, the table lookup mechanism for
Eq.~\eqref{EQ:patternTransitions} is sufficient for not only a regular TM but also a Grand TM that 
contains many TMs and some Universal TMs.

{\bf Scaffolding} Scaffolding means simple skills learned early assist the learning of complex skills later \cite{Vygotsky62,Wood76}.  Imagine that while the automaton grows from ``embryo''  to ``adult'', such meanings become increasingly sophisticated and are internalized as clusters in the later $Y$ areas (see Fig.~\ref{FG:SWWnetFunctions3-2}).   These state/action patterns may also be of any complex meanings, e.g.,  ``goals'' and ``intents'' \cite{WengIEEE-IS2014} that are taught/learned in the ``language'' of actions.   In rats goals have been found in the pre-frontal cortex \cite{Ito15}.  Such meanings may entail creativities, as self-generated programs 
through predictions like those in Eq.~\eqref{EQ:patternTransitions}.   Off-task processes \cite{Solgi13a} (i.e., the automaton takes a short break from 
task execution to ``think'') allow generalizations/creativities, through the seemingly-rigid 
pattern prediction in Eq.~\eqref{EQ:patternTransitions}.   

{\bf Computational complexity}  
Assume that the dimension (e.g., number of pixels) of $X$ is $\alpha$.  Each component 
of $X$ has $10$ possible (e.g., color) values.   Then, there are $c=10^\alpha$ possible $X$ patterns. 

Let the $Z$ area has $\beta$ concept zones (4 in Fig.~\ref{FG:SWWnetFunctions3-2}), where
each zone has $10$ concept values.  There are $r=10^\beta$ possible $Z$ patterns.  

Then, there are $rc=10^\alpha 10^\beta=10^{\alpha+\beta}$ possible patterns in $(\z, \x)$, exponential in $\alpha$ and $\beta$.  For example, when $\alpha=640\times480$ (pixels) and $\beta = 4$ (zones), the transition table for FA already requires $10^{1228800}$ entries, $10^{1228789}$ times more than the number of neurons in a human brain! Note, the table is sparse as
many entries are not observed but could appear in ``life''.

In contrast, the control DN uses a large (e.g., $n=10^{11}$ for a human brain) but constant number $n$ of $Y$ neurons to interpolate among
observed patterns for dealing with exponentially many $rc=10^{\alpha+\beta}$ patterns.
It may use $n\beta$ bottom-up weights of $Z$ to interpolate among observed 
$Y$ patterns for exponentially many possible $Z$ patterns.  
Similarly, it uses $n\alpha$ top-down weights of $X$ to interpolate among observed 
$Y$ patterns for exponentially many possible $X$ patterns.  

Namely, the update for each DN takes
a (large) constant amount of time, so DN has a linear time complexity $O(nt)$ in $t$ while running
in real time $t$ and $n$ is a large constant.  

The GENISAMA TM uses a constant resource to 
optimally (maximum likelihood) interpolate a potentially exponential and unbounded number of observed 
patterns of $(\z, \x)$, conditioned on $k$ in top-$k$ competition, the network size, and training \cite{WengIJIS15}.  For this highly nonlinear optimization problem, local minima may take place in the given $k$ value, the given network size (larger is always better as over-fitting is not a problem with 
the nearest neighbor matching), and the given teaching experience (e.g., teaching complex ideas earlier instead of simple ones earlier). 

This is not a solution to the P=NP? problem \cite{Hopcroft06,Martin11}.  But it suggests that 
if each NP problem is investigated in terms of original patterns, not symbolic,  (e.g., Euclidean space 
\cite{Arora98} as $X$ and learned skills as $Z$), fast and approximate solutions to some NP problems might be available.   

{\bf Proof sketch for Theorem~\ref{TM:Optimal}:} The two properties 1) through 2) have been constructively proved as Theorems 1 through 3 in \cite{WengIJIS15} for the DN to learn from any FA.  But here the teacher is a TM whose control is an Attentive Agent FA according to Theorem~\ref{TM:AttentiveFA}.  We fill this gap.    From the condition that  the patterns from the teacher TM are grounded, the supervision from the teacher TM are the attended pattern patches.  All the new proof needs to do is to replace, everywhere, the attended patch for the monolithic vectors $(\z, \x)$ in the original proofs of \cite{WengIJIS15} (whose main ideas are explained above).  From Theorems 1 through 3 in \cite{WengIJIS15}, the DN learns the pattern-version of the TM transition table with the above properties.  This ends the proof. 

{\bf Proof sketch for Theorem~\ref{TM:auto-program}:}  According to \cite{Hopcroft06,Martin11}, a universal TM $T_u$  corresponds to a subset of transitions in $G$.  It enables the machine to read some $T_i$ and apply the $T_i$ on data in the environment (i.e., tape for TM but the real-world for GENISAMA TM). 
$T_u$ treats some information in the environment as rules and others as data.  However, the mechanism of G table lookup is 
{\em independent} with, and sufficient for, any $T_i$'s and $T_u$ in $G$, as well as  
sharing skills across  $T_i$'s and $T_u$ within the G.   {\em Auto-programming} for {\em general purposes} then corresponds to learning and executing the G,  here ``programming'' because of $T_i$'s, ``general-purpose'' because of $T_u$, and ``auto'' because of sharing transitions among $T_i$'s and $T_u$.    (See Table~\ref{TB:2Tasks} for a sharing example where any of the two 
sub-machines can be replaced by a $T_u$. )  This ends the proof.

Unlike a traditional TM, Theorem~\ref{TM:auto-program} allows any practical language of representation, computer languages or natural languages.   In Eq.~\eqref{EQ:symbolTransitions}, symbols have a static representation, e.g., $q_{T\wedge}$ means
$T$ state followed by $\wedge$.  However in Eq.~\eqref{EQ:patternTransitions}, such symbolic meanings are all hidden in patterns. 
Namely, the meanings of patterns, coded by a language, are in the eyes of the physical world, including teachers.  But the human programmer of the DP does not need to know about such languages or tasks, as shown in Fig.~\ref{FG:TM-GSM-GENISAMA}(c).

{\bf Proof sketch for Theorem~\ref{TM:Church}: }  Prove from (a) to (b) to (c) and then back to (a):  Supposing (a), then (b) follows as a special case of (a) because actions in (b) are for a pencil and the real world in (b) is a piece of paper.  Supposing (b), then (c) holds as a special case because the TM tape with symbols as a special case of paper and  a hand can describe computing steps of any TM transitions.  Finally, suppose (c).  From (c) to (a) is accomplished by a GENISAMA augmentation based on DN:  Let the computing engine simulate the TM that computes DN and augment the TM for GENISAMA (e.g., equip real world sensors and  real world effectors).   Theorem~\ref{TM:auto-program} states this GENISAMA TM inside DN can learn and do auto-programing for general purposes which is (a).  This ends the proof. 

We can see that the major enabling technology here is a successful, complete, and provable decoupling between computing and rich meanings of computing.   Therefore, the algorithm below of DN (i.e., computing) is capable of doing automatic programming for any practically meanings.  This paper does not deal with meanings of computing but the companion paper does. 

Next, let us discuss the Developmental Program (DP) for the controller of the GENISAMA TM:

{\bf Algorithm}
\begin{algorithm}[DP]
\label{AL:DP}
Input areas: $X$ and $Z$.  Output areas: $X$ and $Z$.   The dimension and representation of $X$ and $Z$ areas are determined by the sensors and effectors of the species (or from evolution in biology).   They should also be plastic during prenatal development but for simplicity we assume that they are fixed. 
$Y$ is skull-closed (inside the brain), not directly accessible by the outside.  
\begin{enumerate}
\item At time $t=0$, for each area $A$ in $\{X, Y, Z\}$ (i.e., $A=X$, $A=Y$ and $A=Z$) initialize 
its adaptive part $N=(V, G)$ and the response vector $\r$, where $V$ contains all the synaptic weight vectors and $G$ stores all the neuronal ages.   For example, use the generative DN method discussed below. 
\item At time $t=1, 2, ... $, for each $A$ in $\{X, Y, Z\}$ repeat: 
\begin{enumerate}
\item Every area $A$ performs mitosis-equivalent if it is needed, using its bottom-up input $\b$, lateral input $\r$, and top-down input $\t$, respectively.   The order from bottom to top is $X$, $Y$, and $Z$.   
$X$ does not have bottom-up input.  $Z$ does not have top-down input.   $X$ does not link with $Z$ directly.   The lateral input $\r$ for each neuron includes responses from other neurons in the same area only.  
\item Every area $A$ computes using a globally uniform form of area function $f$, described below, 
\[
(\r', N') = f(\t, \r, \b, N)
\]
where $\t$ is the top-down input (not present for the $Z$ area);
$\b$ the bottom-up input (not present for the $X$ area); 
$\r$ and $\r'$ are area $A$'s old and new response vectors, respectively; and $N$ and $N'$ are the adaptive parts of area $A$, before and after the area update, respectively.    To avoid iterations, lateral  inhibitions that use $A$ to $A$ connection are modeled by top-k competition in Hebbian-like learning below. 
  
\item As asynchronous computation, every area $A$ in $\{X, Y, Z\}$ replaces: $N\leftarrow N'$ and $\r \leftarrow \r'$.
\end{enumerate}
\end{enumerate}
\end{algorithm}
The DN must update at least  twice for the effects of each new signal pattern in $X$ and $Z$, respectively, to go through one update in $Y$ and then one update in $Z$ to appear in $X$ and $Z$.

If $X$ is a sensory area, $\x \in X$ is always supervised.   
The $\z \in Z$ is supervised only when the 
teacher chooses to.   Otherwise, $\z$ gives (predicts) motor output. 

The area function $f$ is based on the theory of Lobe Component Analysis (LCA) \cite{WengLCA09}, a model for self-organization by a neural area.  

Each area neuron with weight $\v=(\v_t, \v_b)$ (both only when exists) in area $A$ has an input vector 
$\p = (\t, \b)$ properly trimmed and 
weighted by synaptic maintenance discussed below.  Its pre-response vector is the sum (or product):
\begin{equation}
\label{EQ:pre-action}
r (\t, \b \;|\; \v_t, \v_b) =  v_t + v_b = \dot{\v_t} \cdot \dot{\t} + \dot{\v_b} \cdot \dot{\b} 
\end{equation}
which measures the degree of match between 
the directions of $\v$ and $\p$, both normalized.  Area $X$ does not have bottom-up part and
area $Z$ does not have the top-down part. 

To simulate lateral inhibitions (winner-take-all) within each area $A$, only top $k$ winners among the $n$ competing neurons fire.  Considering $k=1$, the winner neuron $j$ is identified by:
\begin{equation}
j= \underset{1\le i \le n}{\operatorname{argmax}} \{\dot{\v}_i \cdot \dot{\p}\}.
\label{EQ:jmax}
\end{equation}
Only the single winner fires with response value 
$r'_j=1$ and all other neurons in $A$ do not fire.  The response value $r'_j$ approximates the 
probability for $\dot{\p} $ to fall into the Voronoi region of its $\dot{\v}_j$ where the ``nearness'' 
is $\dot{\v}_j \cdot \dot{\p}$.   
  
All the connections in a DN are learned incrementally 
based on Hebbian learning \cite{Kandel00,Bi01} --- cofiring of the pre-synaptic activity $\dot{\p}$ and the
post-synaptic activity $r'$ of the firing neuron.   If the pre-synaptic end and the post-synaptic
end fire together, the synaptic vector of the neuron has a synapse gain $r'\dot{\p}$.  
Other non-firing neurons do not modify their memory.   
When a neuron $j$ fires, its firing age is incremented $n_j \leftarrow n_j+1$ and then
its synapse vector is updated by a Hebbian-like mechanism:  
\begin{equation}
\v_j \leftarrow w_1(n_j) \v_j + w_2(n_j) r'_j \dot{\p}
\label{EQ:learn}
\end{equation}
where $w_2 (n_j)$ is the learning rate depending on the firing age (counts) $n_j $ of the neuron $j$ and $w_1(n_j)$ is the retention rate with $w_1(n_j) + w_2(n_j) \equiv 1$.  
Note that a component in the gain vector $r'_j \dot{\p}$ is zero if the corresponding component in $\dot{\p}$ is zero.  

The simplest version for $w_2(n_j)$ is $w_2(n_j) = 1/ n_j$.  If the neuron $j$ fires at time $t_i$, $r'_j=1$:
\begin{equation}
 \v_j ^{(i)} = \frac{i-1}{i} \v_j^{(i-1)} + \frac{1}{i} 1 \dot{\p}(t_{i}), \; i = 1, 2, ... , n_j,
 \label{EQ:Hebbian}
\end{equation}
 where $t_i$ is the firing time (not the real time $t=1, 2, 3, ... $) of the post-synaptic neuron $j$.  Gene
 expressions appear to be involved at different times of memory formation \cite{Cho15}.
The above is the recursive way of computing the equally weighted batch average of experience $\dot{\p}(t_i)$:
\begin{equation}
 \v_j ^{(n_j)} = \frac{1}{n_j} \sum_{i=1}^{n_j} \dot{\p}(t_i)
\label{EQ:sum} 
 \end{equation}
 
 In a motivated system, aversive stimuli (e.g., pain) and appetitive stimuli (e.g., pleasure) 
 increase the learning rate, which corresponds to increasing the relative weight for $\dot{\p}(t_i)$ in Eq.\eqref{EQ:sum} for the $Y$ area so that the experience is better memorized.
 However, their effects on actions in the $Z$ area are different: 
 the former and the later inhibits and excites, respectively,  the pre-response values of the corresponding 
 firing neurons in $Z$. 
 
 The initial condition is as follows.   The smallest $n_j$ in Eq.~\eqref{EQ:learn}
   is 1 since $n_j=0$ after initialization.  When $n_j=1$, the initial value of $\v_j$ on the right side of Eq.~\eqref{EQ:learn} is used for pre-response 
competition to find this winner $j$ but the initial value of $\v_j$ does not affect the first-time updated 
 $\v_j$ on the left side since $w_1(1) = 1-1=0$.  

In other words, any initialization of weight vectors will only determine 
who win (i.e., which newly born neurons take the current role) but the initialization will not affect the distribution of weights at all.  In this sense, all random initializations of synaptic weights will work equally well --- all resulting in weight distributions that are computationally equivalent. 
Biologically, we do not care which neurons (in a small 3-D neighborhood) take the specific roles, as long as the distribution of the synaptic weights of these neurons lead to the same computational effect.

If DN learns an Attentive Agent FA as the control of  any TM in which each symbol in $Q$ and $\Gamma$ is represented by a unique natural pattern, the simplest top-1 firing rule is sufficient to be error-free because the number of
sample patterns from the TM is finite.   If DN learns as the control of 
a GENISAMA TM in the real world, the number of samples is infinite.  Then, the limited number of neurons in the control become an optimal representation of the observed probability distribution in the input-state/action space and the top-$k$, $k>1$, firing neurons serve as voting of a committee with 
dynamically changing $k$-member, where the composition of the committee is the top-$k$ best-fit experts.  The number $k$ should be a dynamic number but when fixed it is a conditional parameter of the optimality. 

The ``in-place'' Hebbian learning biologically observed \cite{Kandel91,Montague95,Bi01,Dan06} allows 
each neuron to learn in its own place using its pre-synaptic and post-synaptic activities.  It does not require 
a central controller that is ``aware'' of how to replicate weights across corresponding neurons.  
Convolution is more restricted than the in-place Hebbian learning DN uses here because pattern shifts in convolution only deal with location invariance but not other invariance (e.g., type invariance in location output).   The max-pooling technique originally designed for
convolution for reducing spatial resolution \cite{WengIJCNN92} lead to gaps of ``blind'' locations as shown in \cite{WengCresIJCV97}.  The in-place Hebbian learning here dynamically tolerates shape 
distortion by taking inputs from not only $X$ but also early and later $Y$ neurons.   Early $Y$ neurons 
detect smaller object patches (e.g., head, torso, and limbs for human body detection in this neuron); 
later neurons detect action features that consist of multiple muscle neurons (e.g., action bundles, like
syllables in vocal pronunciation).   Only statistically highly correlated $Y$ neurons will fire and be linked with this neuron because of the synaptic maintenance explained below. 

All the $Z$ neurons may be supervised to fire according to the binary code of $Z(t_i)$.   Consider 
a $Z$ subarea, where each subarea represents a concept (e.g., where, what, or scale) in which only one neuron fires to represent the $i$-th value of the concept. 
For simplicity, consider top-1 firing in the $Y$ area.  Because there is only one $Y$ neuron firing with value 1 at any time
and all other $Y$ neurons respond with value 0, the input to $Z$ is
$\dot \p = \dot \y = \y$.  We can see that the $Z$ neuron $i$ has weight vector 
$\v=(v_1, v_2, ..., v_c)$ in which $v_j$ is the accumulated frequency  $f_j/a_i$ for $Y$ neuron $j$ to fire right before 
the $Z$ neuron $i$ fires, $f_j$ is the number of firings of $Y$ neuron $j$, and $a_i$ is the firing age of $Z$ neuron $i$:
\[
\v = \left( \frac{f_1}{a_i},  \frac{f_2}{a_i} , ... ,  \frac{f_c}{a_i}\right), \mbox{ with } \frac{f_1}{a_i} + \frac{f_2}{a_i} +  ...  + \frac{f_c}{a_i} = 1.
\]
Therefore, as long as the pre-action value of a $Z$ neuron is positive, the $Z$ neuron fires with value 1.
Other $Z$ neurons do not fire.   We can see that the DN prediction of $Z$ firing pattern is always perfect,
as long as DN has observed the transition $(q, \sigma)$ from the FA and has been supervised on its $Z$ for $q'=\delta (q, \sigma)$ when the transition $(q, \sigma)$ is observed for the first time.  No supervision is necessary later for the same transition $(q, \sigma)$. 

The prediction for $X$ is similar to that for $Z$, if the $X$ patterns are binary.   Unlike $Z$, $X$ prediction is not always perfect because FA states are defined for producing the required symbols $q$, but not meant to predict $X$ perfectly.  

Next, let us discuss how the task information is imbedded in a Grand FA, or the control of a 
Grand TM.   Suppose the task information is coded by dedicated neurons, although any action patterns associated with $S_i$ can serve as task context. 
In symbols, each state has
two components $q=(q', q'')$ where $q'$ is the task context, and $q''$ is a state within the task.   For simplicity, consider task 3: count whether the number of inputs of $T, F, \wedge$ is even $q''=q_e$ or odd $q''=q_o$.   Let $s_1\in S_1$ and $s_3\in S_3$ be the sensory stimuli for tasks 1 and 3, respectively, but the default is task 1.   The Grand transition table is shown in Table~\ref{TB:2Tasks}.  

\begin{table}[tb]
\caption{Grand Transition Table for Tasks 1 and 3}
\label{TB:2Tasks}
\begin{center}
\begin{tabular}{|cc|lccccc|}
\hline
\multicolumn{2}{|c|}{$\delta(q,\sigma)$}  & Input $\sigma$ &  $s_1$	& $s_3$ & $T $  &  $F$ &$ \wedge$   \\
 \cline{1-2}
State $(q', q'')=q$ & State pattern $\z$                      & Input pattern $\x$ & 101 &   111 &  010                    & 011              & 100   \\
 \hline
$(q_1, q_0)$ & 01001	& & $(q_1, q_0)$ & $(q_3, q_e)$ & $(q_1, q_{T}) $	& $(q_1, q_{F})$   	&    $(q_1, q_{-})$    \\
$(q_1, q_{T})$ & 01010	& & $(q_1, q_{T})$ & $(q_3, q_e)$ &  $(q_1, q_{-})$    		& $(q_1, q_{-})$    		& $(q_1,  q_{T\wedge} )$ \\
$(q_1, q_{F})$ &  01011	&& $(q_1, q_{F})$ & $(q_3, q_e)$ &  $(q_1, q_{-})$    		& $(q_1, q_{-})$   		& $(q_1, q_{F\wedge} )$ \\
$(q_1,  q_{T\wedge})$ & 01100 &  & $(q_1, q_{T\wedge})$ & $(q_3, q_e)$ & $(q_1, q_{T})$ & $(q_1, q_{F})$ &  $(q_1, q_{-})$ \\
$(q_1, q_{F\wedge})$ & 01101 & & $(q_1, q_{F\wedge})$ & $(q_3, q_e)$ &  $(q_1, q_{F})$ & $(q_1, q_{F})$ & $(q_1, q_{-})$  \\
$(q_1, q_{-})$ & 01110 	& & $(q_1, q_{-})$ & $(q_3, q_e)$ & $(q_1, q_{-})$ 				&$(q_1, q_{-})$ 			&	$(q_1, q_{-})$ \\
\hline
$(q_3, q_e)$  & 11000	& & $(q_1, q_0)$ & $(q_3, q_e)$ & $(q_3, q_o)$ & 	$(q_3, q_o)$ &  $(q_3, q_o)$ \\
$(q_3, q_o)$  & 11001	& & $(q_1, q_0)$ & $(q_3, q_o)$ & $(q_3, q_e)$ & 	$(q_3, q_e)$ &  $(q_3, q_e)$ \\
\hline
\end{tabular}
\end{center}
\label{TB:ANDMix}
\end{table}%

Table~\ref{TB:2TasksPattern} gives the pattern-only transition table.  We can see that the mechanism of table lookup is independent of the meanings of the machines inside.  Namely, the GENISAMA TM's control DN is for general purposes. 

\begin{table}[thb]
\caption{Pattern-Only Grand Transition Table for Tasks 1 and 3}
\label{TB:2TasksPattern}
\begin{center}
\begin{tabular}{|c|lccccc|}
\hline
State pattern $\z$                      & Input pattern $\x$ & 101 &   111 &  010                    & 011              & 100   \\
 \hline
 01001	& &  01001 & 11000 & 01010	&  01011   	&    01110    \\
 01010	& & 01010 & 11000 &  01110    		& 01110    		& 01100 \\
  01011	&&  01011 & 11000 &  01110    		& 01110   		& 01101 \\
 01100 &  & 01100 & 11000 & 01010 &  01011 &  01110 \\
 01101 & & 01101 & 11000 &   01011 &  01011 & 01110  \\
01110 	& & 01110 & 11000 & 01110 				&01110 			&	01110 \\
 11000	& & 01001 & 11000 & 11001 & 	11001 &  11001 \\
 11001	& & 01001 & 11001 & 11000 & 	11000 &  11000 \\
\hline
\end{tabular}
\end{center}
\label{TB:ANDPattern}
\end{table}%

An experienced teacher would teach simpler skills first so that they facilitate the learning of more complex  skills later---a process known as scaffolding \cite{Vygotsky62,Wood76}.   In particular, the Grand Teacher TM should teach $T_i$ before teach $T_u$ because the latter calls the former.  

The generality of the GENISAMA TM formulation instantiated by the above Table~\ref{TB:2TasksPattern} example casts light on the popular nature-nurture debate \cite{McClelland94}.  In the formulation, the genome-like (largely nature) DP is body-specific (which may include 
body-specific inborn behaviors) but task-nonspecific.
 The DP enables table lookup using exclusively 
patterns like Table~\ref{TB:2TasksPattern} (which may include tasks and discoveries
that the parents never knew).  The contents inside the table are task-specific (largely nurture), emerging automatically from the interactions among the external world (sensed and effected environment) through the sensors and effectors, the internal world (inside DN), and the DP.   Nature and nurture are inseparable but their roles are clear in the model. 

{\bf Motivation} is very rich.  It has two major aspects (a) and (b) in the current DN model.   All reinforcement learning methods other than DN, as far as we know, are for symbolic methods (e.g., Q-learning \cite{Sutton98,Mnih15}) and are in aspect (a) exclusively.   DN uses concepts (e.g., important events) instead of the rigid time-discount in Q-learning to avoid the failure of far goals.

(a) Pain avoidance and pleasure seeking to speed up learning important events.  Signals from pain  (aversive) sensors release a special kind of neural transmitters (e.g., serotonin \cite{Daw02})  that diffuse into all neurons that suppress $Z$ firing neurons but speed up the learning rates of the firing $Y$ neurons.    Signals from sweet (appetitive) sensors release a special kind of neural transmitters (e.g., dopamine \cite{Kakade02})  that diffuse into all neurons that excite $Z$ firing neurons but also speed up the learning rates of the firing $Y$ neurons.   Higher pains (e.g., 
loss of loved ones and jealousy) and higher pleasure (e.g., praises and respects) develop at later ages from lower pains and pleasures, respectively. 

(b) Synaptic maintenance ---grow and trim the spines of synapses \cite{Wang11,Guo15} --- to 
segment object/event and motivate curiosity.   Each synapse incrementally estimates the average error $\beta$ between the pre-synaptic signal and
the synaptic conductance (weight), represented by a kind of neural transmitter (e.g.,  acetylcholine \cite{Yu05}). 
Each neuron estimates the average deviation $\bar\beta$ as the average across
all its synapses.   
The ratio $\beta/\bar\beta$ is the novelty represented by a kind of neural transmitters (e.g., norepinephrine, \cite{Yu05}) at each synapse.  The synaptogenic factor $f(\beta, \bar\beta)$ at each synaptic spine
and full synapse enables the spine to grow if the ratio is low (1.0 as default) and
to shrink if the ratio is high (1.5 as default).  Each area $X$, $Y$, and $Z$ 
has a prenatal (default) hierarchy of subareas and subsubareas (e.g. Brodmann areas and its subareas for $Y$) that 
continuously adapt postnatally.   Each area, subarea, subsubarea, has its own synaptogenic factor.   This network of synaptogenic factors 
dynamically organize the complex brain network (e.g., \cite{FellemanVanEssen91}).  See Fig.~\ref{FG:SWWnetFunctions3-2}(b) for how a neuron can cut off their direct connections with $Z$ to become 
early areas in the occipital lobe or their direct connections with the $X$ areas to become latter areas inside
the parietal and temporal lobes.   However, we cannot guarantee that such ``cut off'' are 100\% based on the statistics-based wiring theory here. 

Table~\ref{TB:comparison} compares TMs, Universal TMs, Grounded Symbolic Machines, prior 
neural networks, and GENISAMA TMs. 

\begin{table}[htp]
\caption{Different Types of Machines}
\begin{center}
\begin{tabular}{|c|ccccc|}
\hline
Type of Machines & TMs & Universal TMs & Grounded Symbolic Machines & Prior Neural Networks & GENISAMA TMs \\
\hline
Unknown Tasks	 & No		& Yes		& No  & Pattern recognition only			& Yes \\
General purpose & No			& Yes		& No  & No			& Yes\\
Grounded 		& No			&No			& Yes & Yes (can be)		& Yes \\
Auto-program		& No			& No			&No & No			& Yes\\
\hline
\end{tabular}
\end{center}
\label{TB:comparison}
\end{table}%

The experimental results are reported in the companion letter which 
shows how far we are toward fully automatic machine learning --- fully automatic 
programming occurs as short transitions of TM.   However, the theory here shows that the length and the complexity of the learned knowledge are not limited by the methodology.   Thus, strong AI is now 
possible in the presented theory.  

\bibliographystyle{naturemag}
\bibliography{shoslifref}

\section*{Addendum}
\begin{description}
\item This version of the writing was submitted June 13, 2017 to Nature as one of the two joint letters.  The other letter is experimental, about vision, audition, and natural language acquisition using the theory and methods here.   The author would like to thank Drs. Matt Mutka, Jiayu Zhou, Hu Ding, Min Jiang and Mr. Sorrachai Yingchar for reading earlier versions of the manuscript and provided valuable comments that improved the presentation.  Correspondence and requests for materials
should be addressed to J. Weng (weng@cse.msu.edu).
\end{description}


\end{document}